\documentclass{article}

\usepackage{microtype}
\usepackage{graphicx}
\usepackage{multirow}
\usepackage{float}
\usepackage{subfigure}
\usepackage{booktabs} % for professional tables

\usepackage{hyperref}

\usepackage[accepted]{icml2023}

\usepackage{amsmath}
\usepackage{amssymb}
\usepackage{mathtools}
\usepackage{amsthm}

\usepackage[capitalize,noabbrev]{cleveref}

\theoremstyle{plain}

\theoremstyle{definition}

\theoremstyle{remark}

\usepackage[textsize=tiny]{todonotes}

\icmltitlerunning{L3Cube-MahaSent-MD: A Multi-domain Marathi Sentiment Analysis Dataset}

\begin{document}

\twocolumn[
\icmltitle{L3Cube-MahaSent-MD: A Multi-domain Marathi Sentiment Analysis Dataset and Transformer Models}




\begin{icmlauthorlist}
\icmlauthor{Aabha Pingle}{yyy,comp}
\icmlauthor{Aditya Vyawahare}{yyy,comp}
\icmlauthor{Isha Joshi}{yyy,comp}
\icmlauthor{Rahul Tangsali}{yyy,comp}
\icmlauthor{Raviraj Joshi}{xxx,comp}
\end{icmlauthorlist}

\icmlaffiliation{yyy}{Department of Computer Engineering, SCTR'S Pune Institute of Computer Technology, Pune, India}
\icmlaffiliation{comp}{L3Cube, Pune, India}
\icmlaffiliation{xxx}{Indian Institute of Technology Madras, Chennai, India}

\icmlcorrespondingauthor{Aabha Pingle}{aabhapingle@gmail.com}
\icmlcorrespondingauthor{Aditya Vyawahare}{aditya.vyawahare07@gmail.com}
\icmlcorrespondingauthor{Isha Joshi}{joshiishaa@gmail.com}
\icmlcorrespondingauthor{Rahul Tangsali}{rahuul2001@gmail.com}
\icmlcorrespondingauthor{Raviraj Joshi}{ravirajoshi@gmail.com}

\icmlkeywords{Machine Learning, Marathi NLP, Marathi Sentiment Analysis, Marathi Sentiment Datasets, Multi-domain Sentiment Datasets, Low-resource NLP}

\vskip 0.3in
]



\printAffiliationsAndNotice{\icmlEqualContribution} % otherwise use the standard text.

\begin{abstract}
The exploration of sentiment analysis in low-resource languages, such as Marathi, has been limited due to the availability of suitable datasets. In this work, we present L3Cube-MahaSent-MD, a multi-domain Marathi sentiment analysis dataset, with four different domains - movie reviews, general tweets, TV show subtitles, and political tweets. The dataset consists of around 60,000 manually tagged samples covering 3 distinct sentiments - positive, negative, and neutral. We create a sub-dataset for each domain comprising 15k samples. The MahaSent-MD is the first comprehensive multi-domain sentiment analysis dataset within the Indic sentiment landscape.  We fine-tune different monolingual and multilingual BERT models on these datasets and report the best accuracy with the MahaBERT model.  We also present an extensive in-domain and cross-domain analysis thus highlighting the need for low-resource multi-domain datasets. The data and models are available at {\small \url{https://github.com/l3cube-pune/MarathiNLP}} .
\end{abstract}

\section{Introduction}

Sentiment analysis is the process of determining the sentiment or emotional tone expressed in a piece of text, such as a sentence or a group of sentences. It involves analyzing the subjective content to identify whether it conveys positive, negative, or neutral sentiments \cite{pang-etal-2002-thumbs}. Sentiment analysis is a valuable technique in natural language processing (NLP) and finds applications across different domains \cite{blitzer-etal-2007-biographies}. It can be applied to various forms of text data, including tweets, social media posts, customer reviews, surveys, news articles, and more \cite{roccabruna-etal-2022-multi}. Common uses of sentiment analysis include customer feedback analysis, social media monitoring, market research, financial analysis, and more.

Sentiment analysis can be utilized to aggregate sentiments expressed in movie reviews and calculate an overall sentiment score or rating for a movie \citep{maas-etal-2011-learning}. This enables users to quickly grasp the general sentiment towards a movie prior to watching it. Real-time monitoring of tweets \citep{ekbal2020multi} and detection of sentiment trends related to specific topics, events, or hashtags can be achieved through sentiment analysis. This aids businesses and marketers in comprehending public opinion, identifying emerging trends, and adjusting their strategies accordingly. Categorizing the emotional tone of different scenes or segments through sentiment analysis of subtitles is valuable. This can enhance content searchability and retrieval, making it easier for viewers to find specific types of content based on their moods. While these applications have been extensively researched in high-resource languages like English \cite{zhang2018deep}, they are lacking in low-resource languages like Marathi \cite{joshi2022l3cube_mahanlp,kulkarni2022experimental,joshi2022l3cube,lahoti2022survey}.

In this work, we present the L3Cube-MahaSent-MD dataset, an expanded version of the original L3CubeMahaSent \cite{kulkarni-etal-2021-l3cubemahasent} dataset that specifically focuses on sentiment analysis in the Marathi language. This dataset includes sub-datasets in four domains, with three new domains introduced in this work: MahaSent-MR (movie reviews), MahaSent-GT (generic tweets), MahaSent-ST (TV show subtitles), and MahaSent-PT (political tweets). MahaSent-MR consists of Marathi movie reviews obtained by scraping various websites. MahaSent-GT includes a collection of Marathi tweets covering diverse topics, collected from Twitter using advanced techniques. MahaSent-ST consists of translated Marathi subtitles from the popular English TV show "Friends". Lastly, MahaSent-PT represents the original L3Cube-MahaSent dataset, specifically containing political tweets. Each dataset comprises approximately 15,000 examples in native Devanagari script, divided into training, validation, and test sets. The annotation policies employed for all these datasets are thoroughly described. Further, we also present the results of standard transformer-based BERT models on these datasets. The models used to evaluate the accuracies are MuRIL \footnote{\href{https://huggingface.co/google/muril-base-cased}{muril-base-cased}} \cite{Khanuja2021MuRILMR}, mBERT \footnote{\href{https://huggingface.co/bert-base-multilingual-cased}{bert-base-multilingual-cased}} \cite{devlin-etal-2019-bert}, MahaBERT \footnote{\href{https://huggingface.co/l3cube-pune/marathi-bert-v2}{marathi-bert-v2}} \cite{joshi2022l3cube}, and IndicBERT \footnote{\href{https://huggingface.co/ai4bharat/indic-bert}{ai4bharat/indic-bert}} \cite{kakwani-etal-2020-indicnlpsuite}. We also perform cross-domain analysis on these datasets using MahaBERT, the best-performing model. The main contributions of this work are as follows:
\begin{itemize}
\item We present \textbf{L3Cube-MahaSent-MD}, the first comprehensive multi-domain sentiment analysis dataset in Marathi, an Indian language. It comprises approximately 60,000 manually tagged sentences across four different domains, with positive, negative, and neutral labels. Specifically, our work contributes three new domains: movie reviews, generic tweets, and TV show subtitles, encompassing around 45,000 manually tagged sentences.
\item This study introduces sub-datasets for sentiment analysis, namely MahaSent-MR, a dataset for Marathi Movie Reviews, MahaSent-GT, a dataset of Marathi general tweets, and MahaSent-ST, a dataset of Marathi subtitles. These datasets are the first of their kind and address a gap in the existing literature.
\item We evaluate various monolingual and multilingual BERT models like MahaBERT, MuRIL, mBERT, and IndicBERT and demonstrate the superior performance of the monolingual MahaBERT model. Additionally, we release MahaBERT-based Marathi sentiment models for the domains considered in this study. 
\item We also conduct cross-domain analysis to demonstrate that domain-specific models do not generalize well across different domains. Thus, there is a significant need for multi-domain datasets in low-resource languages. We show that a multi-domain MahaBERT model trained on all the domains works competitively with the domain-specific models.
\end{itemize}

The datasets and models are shared publicly\footnote{\href{https://github.com/l3cube-pune/MarathiNLP}{marathi-nlp}}. The individual models for MahaSent-MR\footnote{\href{https://huggingface.co/l3cube-pune/marathi-sentiment-movie-reviews}{marathi-sentiment-movie-reviews}}, MahaSent-GT\footnote{\href{https://huggingface.co/l3cube-pune/marathi-sentiment-tweets}{marathi-sentiment-tweets}}, MahaSent-ST\footnote{\href{https://huggingface.co/l3cube-pune/marathi-sentiment-subtitles}{marathi-sentiment-subtitles}}, MahaSent-PT\footnote{\href{https://huggingface.co/l3cube-pune/marathi-sentiment-political-tweets}{marathi-sentiment-political-tweets}}, and MahaSent-All\footnote{\href{https://huggingface.co/l3cube-pune/marathi-sentiment-md}{marathi-sentiment-md}} are available on Huggingface.

\iffalse
\begin{figure*}
	\begin{center}
		\includegraphics[scale=0.3]{MovieReview.png}
		\caption{Sample sentences from the MahaSent-MR dataset}
	\end{center}
\end{figure*}

\begin{figure*}
	\begin{center}
		\includegraphics[scale=0.3]{Subtitle.png}
		\caption{Sample sentences from the MahaSent-ST dataset}
	\end{center}
\end{figure*}
\begin{figure*}
	\begin{center}
		\includegraphics[scale=0.3]{GeneralTweet.png}
		\caption{Sample sentences from the MahaSent-GT dataset}
	\end{center}
\end{figure*}
\fi
\begin{figure*}
\hfill
\subfigure[MahaSent-MR dataset]{\includegraphics[width=5cm]{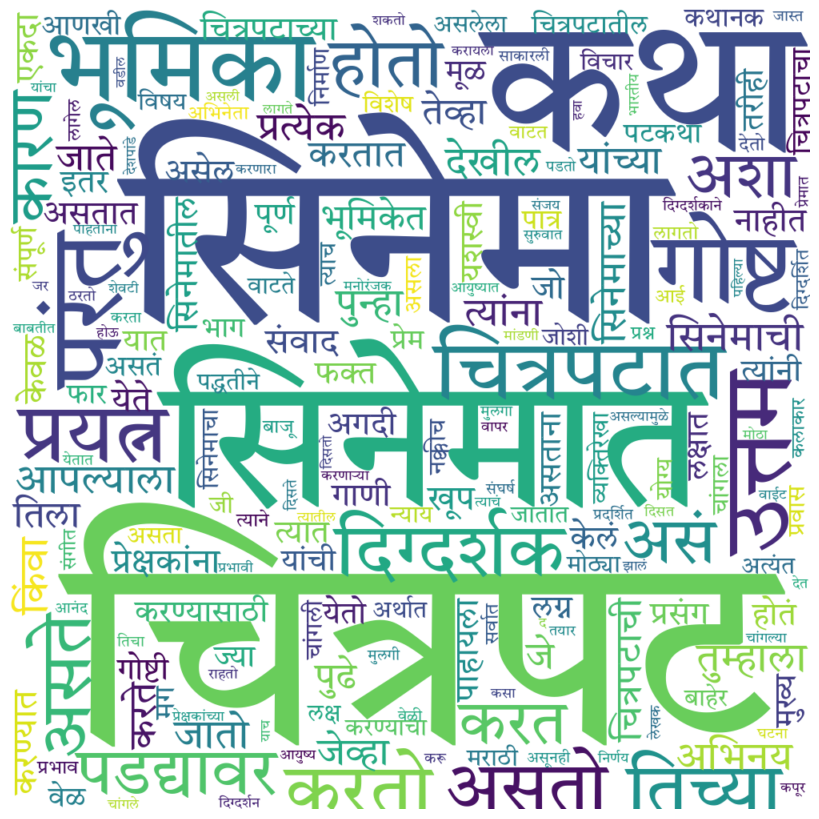}}
\hfill
\subfigure[MahaSent-ST dataset]{\includegraphics[width=5cm]{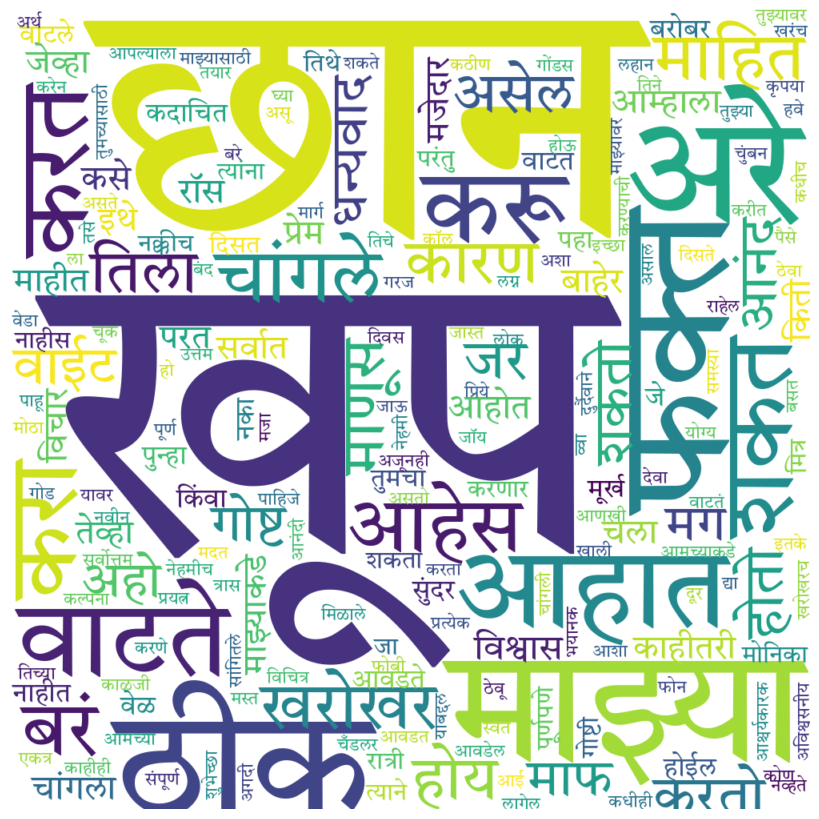}}
\hfill
\subfigure[MahaSent-GT dataset]{\includegraphics[width=5cm]{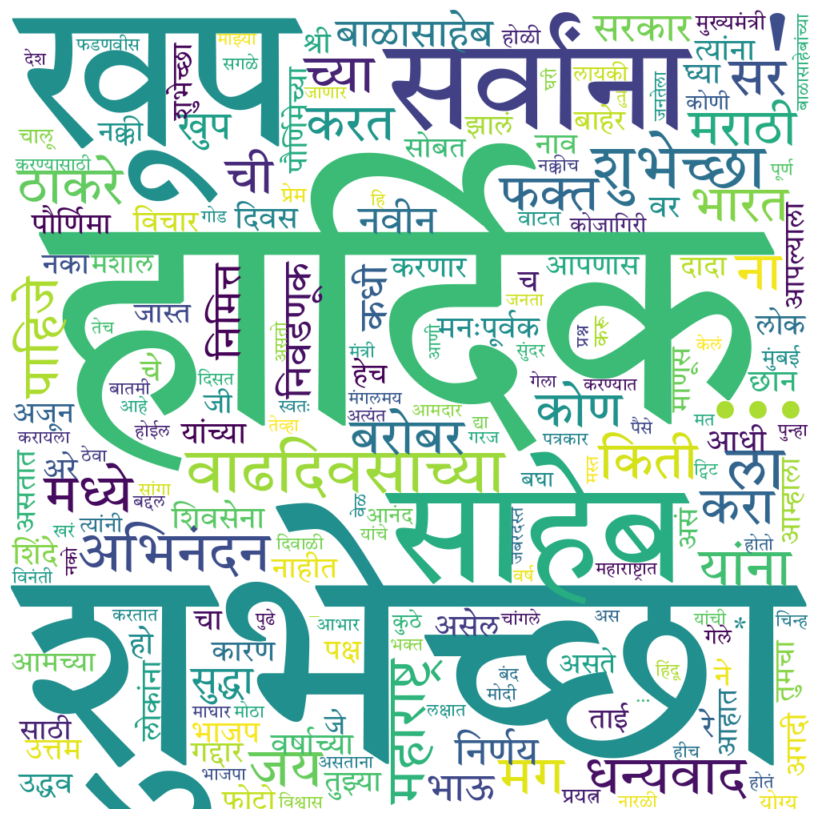}}
\hfill
\caption{Word clouds for the movie reviews, subtitles, and generic tweets domain}
\end{figure*}

\section{Related Work}

In this section, we will discuss multi-domain datasets and low-resource datasets in the field of sentiment analysis.
A vast amount of content is published on the internet every day, which has led to platforms like Twitter gaining significant attention for sentiment analysis tasks \citep{ekbal2020multi}. Additionally, domains such as movie reviews have remained highly popular for sentiment analysis \citep{maas-etal-2011-learning}.

\citep{roccabruna-etal-2022-multi} emphasizes the importance of enhancing the performance of BERT models across multiple sources and domains. The authors conduct an extensive evaluation of BERT-based models using sentiment analysis corpora from various domains and sources. Their research indicates that jointly fine-tuning the model on multi-source and multi-domain corpora yields superior performance compared to fine-tuning it solely on single-source and single-domain settings.

We conducted a thorough review of numerous sentiment analysis datasets available for low-resource languages. One of these datasets is HindiMD, a multi-domain corpus specifically developed for Hindi sentiment analysis \cite{ekbal2022hindimd}. This dataset consists of a total of 9,090 tweets written in Hindi, which is considered a low-resource Indic language. The authors obtained these texts from Twitter by utilizing various keywords for data crawling. Through the creation of multiple baselines using this dataset, they effectively showcased its usefulness. However, it should be noted that this work does not differentiate between different domains and provides a single dataset encompassing all domains. As a result, the usability of the datasets for cross-domain analysis is limited.

\cite{inproceedings} describes their work on cross-lingual sentiment analysis for Indian languages using linked wordnets. The dataset utilized in their study comprised user-written travel destination reviews. The authors conducted several experiments on this dataset, specifically focusing on the Hindi and Marathi languages.

\citep{inproceedings1} utilized a dataset for their own experimentation on sentiment analysis in Marathi. This dataset included Marathi and Hindi documents sourced from social media platforms, encompassing data from chats, tweets, and YouTube comments. Notably, the dataset consisted of text in the transliterated (Romanized) format.

Released in 2021, the L3CubeMahaSent dataset \cite{kulkarni-etal-2021-l3cubemahasent} is a manually labeled dataset comprising Marathi tweets. It consists of approximately 18,378 tweets that were classified into three sentiment classes. This dataset stands as the largest publicly available resource for Marathi sentiment analysis to date. However, it should be noted that the dataset exclusively contains tweets from the political domain. A similar tweet-based Marathi hate speech detection corpus, MahaHate was released in \cite{patil2022l3cube}.
Both MahaSent and MahaHate datasets have also been evaluated in a separate study \cite{velankar2022mono}. A dataset for named entity recognition in Marathi, MahaNER was released in \cite{litake2022l3cube}.

\section{Dataset Collection}

\subsection{MahaSent-MR (Marathi Movie Reviews)}
We identified a number of Marathi movie review websites in native Devanangari script for the MahaSent-MR dataset. The Marathi language movie reviews were scraped from TV9 \footnote{\href{https://www.tv9marathi.com/live-tv}https://www.tv9marathi.com/live-tv}
, eSakal \footnote{\href{https://www.esakal.com/}https://www.esakal.com/}
, Loksatta \footnote{\href{https://www.loksatta.com/}https://www.loksatta.com/}
, ABP Majha \footnote{\href{https://marathi.abplive.com/}https://marathi.abplive.com/} , Lokmat \footnote{\href{https://www.lokmat.com/}https://www.lokmat.com/}, Saamana \footnote{\href{https://www.saamana.com/}https://www.saamana.com/}
 and Maharashtra Times \footnote{\href{https://maharashtratimes.com/}https://maharashtratimes.com/}. 
 %English movie reviews were obtained from The Times of India \footnote{\href{https://timesofindia.indiatimes.com/?from=mdr}https://timesofindia.indiatimes.com/?from=mdr} and Cinestaan \footnote{\href{https://www.cinestaan.com/}https://www.cinestaan.com/} and translated to Marathi using the deep-translator tool \footnote{\href{https://pypi.org/project/deep-translator/}https://pypi.org/project/deep-translator/}. 
 % There are 2062 sentences which were scraped from English language reviews that were translated. 324 are positive, 1738 are negative
 The movie reviews were scraped using the Requests module  \footnote{\href{https://pypi.org/project/requests/}https://pypi.org/project/requests/} in Python. The content of the articles was extracted from the webpages using the re (regular expressions) \footnote{\href{https://docs.python.org/3/library/re.html}https://docs.python.org/3/library/re.html} Python module. The movie review articles obtained were then split into individual sentences using the period symbol (“.”) as the delimiter. All special characters were removed from the text.

\subsection{MahaSent-ST (Marathi Subtitles)}
The MahaSent-ST dataset includes English subtitles that have been translated into Marathi using the deep-translator tool \footnote{\href{https://pypi.org/project/deep-translator/}https://pypi.org/project/deep-translator/}. All translations we manually verified and problematic generations were discarded. Subtitles from the situational comedy TV show F.R.I.E.N.D.S. have been used to create the dataset. The dataset was created using subtitles from seasons 1 through 4. The subtitle files had the .sub and .srt file extensions and were converted to text (.txt) files. The text file subtitles were then divided into individual sentences in the dataset. Special characters have not been removed from the dataset.

\subsection{MahaSent-GT (Marathi Generic Tweets)}
The MahaSent-GT dataset contains tweets in the Marathi language (native script). To scrape tweets, we utilized the Python tools twint \footnote{\href{https://pypi.org/project/twint/}https://pypi.org/project/twint/}  and snscrape \footnote{\href{https://github.com/JustAnotherArchivist/snscrape}https://github.com/JustAnotherArchivist/snscrape}. In order to ensure that the dataset contained generic tweets from wide range of topics, we scraped tweets using non-specific (stopwords in Marathi) keywords. We restricted the length of tweets to 15 words or less. 

%When scraping additional negative or positive tweets using certain keywords, we limited the number of tweets chosen from a given domain to minimize over-representation from any specific domain. We used positive Marathi words (such as "congratulations") paired with emoticons used in a positive context to obtain more positive tweets. The tweets were then combined, and duplicates were removed. 
\subsection{MahaSent-PT (Marathi Political Tweets)}
The original MahaSent dataset \citep{kulkarni-etal-2021-l3cubemahasent}, referred to as MahaSent-PT in this work, contains tweets regarding current affairs. It features tweets in Marathi from political figures and activists presenting a variety of thoughts and perspectives. The dataset was manually categorized into three categories: negative, positive, and neutral. In the publicly available version of the dataset, hashtags, mentions, symbols and occasional English words have been retained. The tweets were scraped using the twint library.

\begin{figure*}
\hfill

\subfigure[MahaSent-MR dataset]{\includegraphics[scale=0.3]{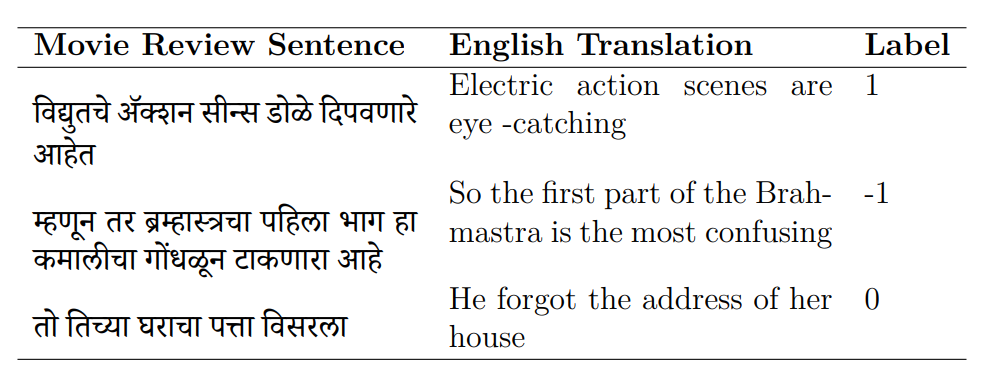}}
\hfill
\subfigure[MahaSent-GT dataset]{\includegraphics[scale=0.3]{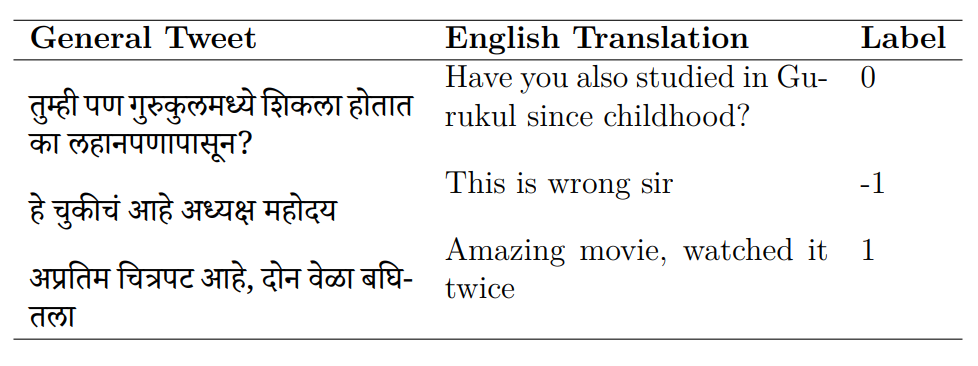}}
\hfill
\begin{center}
\centering
\subfigure[MahaSent-ST dataset]{\includegraphics[scale=0.3]{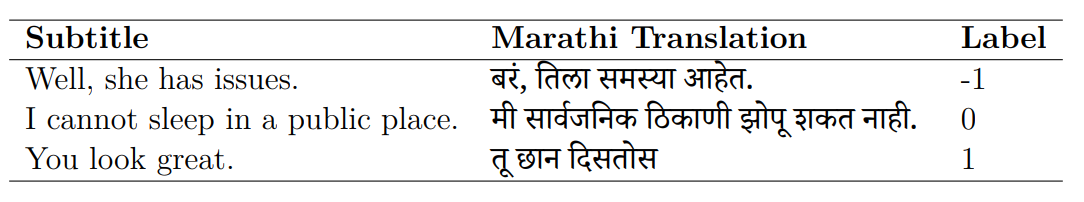}}
\end{center}

\hfill
\caption{Sample sentences from the movie reviews, subtitles, and generic tweets domain}
\end{figure*}

\section{Dataset Annotation}

The sentences in the datasets were classified into three categories: positive (1), negative (-1) and neutral (0). All the datasets were manually annotated by four native Marathi speakers who are proficient in reading and writing Marathi. The Cohen's Kappa \cite{cohen1960coefficient} for the annotators is 0.86. The scraped setences were initially annotated using existing out-of-domain sentiment models in-order to aid the manual process. The Marathi MahaSent-MR and MahaSent-GT sentences were tagged (as positive, negative or neutral) using the MarathiSentiment model \footnote{\href{https://huggingface.co/l3cube-pune/MarathiSentiment}https://huggingface.co/l3cube-pune/MarathiSentiment}. Similarly, the non-traslated English sentences from the MahaSent-ST dataset were tagged using VADER (Valence Aware Dictionary and sEntiment Reasoner)\citep{hutto2014vader} from the NLTK \footnote{\href{https://www.nltk.org/}https://www.nltk.org/} library. The sentences were then sorted according to the tags, and the negative and positive sentences were annotated first. This was done to expedite the annotation process, establish a balanced dataset, and prevent the annotation of excessive neutral sentences. The initial out-domain models utilized for pseudo-labeling achieved an accuracy of 60-70\% on the final datasets, highlighting the importance of manual annotation for new domains.

A set of guidelines were employed for the annotation of the datasets to ensure consistency in labeling. For the MahaSent-MR dataset, a sentence was labeled negative if a reviewer appeared to complain about an aspect of the film or reported a negative incident in the film. Similarly, each sentence that featured praise for the film or a description of some favorable developments in the story was labeled as a positive sentence. Sentences that simply provided facts about the movie or its events were considered neutral sentences.

One-word or small subtitles in the MahaSent-ST dataset were categorized based on the sentiment of the words present, regardless of context. "Yes", for example, was labeled as a positive subtitle. Grammatical errors in the subtitle translations have been rectified by the annotators. Incorrect but grammatically sound translations have not been altered. 

Sarcastic tweets annotated for the MahaSent-GT dataset were marked negative. If the context of a tweet could not be ascertained, it was labeled as neutral. While tagging, English hashtags and emoticons were ignored, but Marathi hashtags were taken into account to determine the sentiment. Tweets offering condolences or expressing regret were categorized as negative.

Figures 2(a), 2(b) and 2(c) illustrate some sample sentences from the MahaSent-MR, MahaSent-GT and MahaSent-ST datasets, respectively.

\begin{table*}
\centering
\caption{\label{Dataset_Statistics}Dataset Statistics}
\begin{tabular}{lccccc}
\hline
\textbf{Dataset} & \textbf{Total Samples} & \textbf{Train} & \textbf{Valid} & \textbf{Test} & \textbf{Average Sentence Length}\\
\hline
MahaSent-PT (political tweets) & 15864 & 12114 & 1500 & 2250 & 26.8658 \\
MahaSent-GT (generic tweets) & 15000 & 12000 & 1500 & 1500 & 10.4528 \\
MahaSent-ST (subtitles) & 15000 & 12000 & 1500 & 1500 & 6.3753 \\
MahaSent-MR (movie review) & 15000 & 12000 & 1500 & 1500 & 12.6323 \\
MahaSent-All (all) & 60864 & 48114 & 6000 & 6750 & 14.2630\\
\hline
\end{tabular}

\end{table*}

\section{Dataset Statistics}
\indent The updated version of L3CubeMahaSent consists of a total of 15000 records for MahaSent-MR, MahaSent-GT, and MahaSent-ST each. The MahaSent-MR domain consists of movie reviews, MahaSent-GT consists of general tweets, and MahaSent-ST comprises of TV show subtitles. Each of these domains has been annotated for sentiment analysis in the Marathi language. The dataset has been curated to ensure that the classes are balanced by randomly selecting an equal number of tweets for each class.

\indent The MahaSent-MR, MahaSent-GT, and MahaSent-ST datasets have been split into train, test, and validation sets, where 12000 records of the data were used for training and 1500 records each were used for validation and testing. 

\indent We also create a single corpus by merging all four datasets (MahaSent-PT, MahaSent-MR, MahaSent-GT, and MahaSent-ST) together and maintaining an equal number of labels of each sentiment in the training, validation, and test sets. There are 48114 total examples in the training set containing 16038 examples of each sentiment, 6000 total examples in the validation set containing 2000 examples of each sentiment, and 6750 total examples in the test set containing 2250 examples of each sentiment. To further ensure the quality of the dataset, the commonly occurring words in each class have been visualized in the form of word clouds, which can be seen in Figures 1(a), 1(b), and 1(c). Table \ref{Dataset_Statistics} gives a detailed view of the statistics of each of the datasets that have been used. The L3Cube-MahaSent-MD dataset is expected to facilitate sentiment analysis research in the Marathi language across multiple domains.

\section{Baseline Models}
\subsection{mBERT} 
mBERT, also known as multilingual BERT, is a BERT-based model trained on and applicable to 104 languages \cite{devlin-etal-2019-bert}. It can be effectively utilized for downstream sentiment analysis tasks and was trained using the masked language modeling (MLM) and next sentence prediction (NSP) objectives.

To conduct sentiment analysis on Marathi text using mBERT, one can fine-tune the model on a Marathi sentiment analysis dataset. In this process, a new classification layer can be trained on top of mBERT. This new layer maps the representations learned by mBERT to the appropriate sentiment labels.
\subsection{IndicBERT}
IndicBERT is a language model based on the ALBERT architecture \cite{Lan2020ALBERT:} and has been trained on a substantial corpus covering 12 major Indian languages: Assamese, Bengali, English, Gujarati, Hindi, Kannada, Malayalam, Marathi, Oriya, Punjabi, Tamil, and Telugu. The training data for IndicBERT was sourced from the IndicCorp dataset \cite{kakwani-etal-2020-indicnlpsuite} and utilized joint training, which enables effective usage for underrepresented languages. In comparison to the XLM-R \cite{conneau-etal-2020-unsupervised} and mBERT models, IndicBERT generally demonstrates superior performance. Two variations of the model are available: IndicBERT base, trained on 12 million tokens, and IndicBERT large, trained on 18 million tokens.
\subsection{MuRIL}
MuRIL is a language model specifically developed for Indian languages and is trained exclusively on a substantial amount of Indian text data \cite{Khanuja2021MuRILMR}. To provide supervised cross-lingual signals during training, translated and transliterated document pairings are incorporated into the training set. This approach enhances MuRIL's ability to capture the nuances of Indian languages and effectively handle transliterated input. The performance of MuRIL was evaluated using the XTREME benchmark \cite{hu2020xtreme}, which involves challenging cross-lingual evaluation tasks. Across the board, MuRIL outperformed multilingual BERT (mBERT) based on the evaluation results. Furthermore, transliterated test sets were utilized to assess the model's proficiency in handling such data.
\subsection{MahaBERT}
MahaBERT is a multilingual BERT model that has been fine-tuned on L3Cube-MahaCorpus, as well as other publicly available Marathi monolingual datasets \cite{joshi2022l3cube}. It was trained using the masked language modeling objective and trained on a corpus comprising 752M tokens.

\begin{table*}
\centering
\caption{\label{Traning Accuracies}Classification accuracies for the datasets using different BERT models}
\begin{tabular}{lcccc}
\hline
\textbf{Model} & \textbf{MahaSent-PT} & \textbf{MahaSent-GT} & \textbf{MahaSent-ST} & \textbf{MahaSent-MR} \\
\hline
mBERT & 80.66 & 70.07 & 75.26 & 70.53 \\
IndicBERT & 84.13 & 76.06 & 77.00 & 74.13 \\
MuRIL & 84.30 & 77.13 & 78.73 & 78.00 \\
MahaBERT & \textbf{84.90} & \textbf{78.80} & 79.07 & \textbf{78.53} \\
MahaBERT-All & 83.95 & 77.86 & \textbf{79.20} & 77.73 \\
\hline
\end{tabular}

\end{table*}

\begin{table*}
\centering
\caption{\label{Cross-Domain Analysis}
Cross-domain analysis of models trained on different datasets}
\begin{tabular}{lccccc}
%\begin{tabular}{ p{0.75in} | p{0.75in}p{0.75in}p{0.75in}p{0.75in}p{0.75in}}
% \begin{tabular}{p{1in}p{0.89in}p{0.89in}p{0.89in}p{0.89in}p{0.89in}}
\hline
\textbf{Model} & 
\hspace{-0.15cm}\textbf{MahaSent-PT} & 
\hspace{-0.15cm}\textbf{MahaSent-GT} &
\hspace{-0.15cm}\textbf{MahaSent-ST} &
\hspace{-0.15cm}\textbf{MahaSent-MR} &
\hspace{-0.15cm}\textbf{MahaSent-All} \\
\hline
MahaBERT-PT  & \hspace{-0.2cm} \textbf{84.90} & \hspace{-0.2cm} 69.87 & \hspace{-0.2cm} 60.60 & \hspace{-0.2cm} 62.93 & \hspace{-0.2cm} 70.80\\
MahaBERT-GT  & \hspace{-0.2cm} 70.40 & \hspace{-0.2cm} \textbf{78.80} & \hspace{-0.2cm} 67.73 & \hspace{-0.2cm} 67.13 & \hspace{-0.2cm} 70.91\\
MahaBERT-ST & \hspace{-0.2cm} 75.24 & \hspace{-0.2cm} 72.80 & \hspace{-0.2cm} 79.07 & \hspace{-0.2cm} 70.33 & \hspace{-0.2cm} 74.46\\
MahaBERT-MR  & \hspace{-0.2cm} 73.33 & \hspace{-0.2cm} 66.46 & \hspace{-0.2cm} 61.73 & \hspace{-0.2cm} \textbf{78.53} & \hspace{-0.2cm} 70.42\\
MahaBERT-All & \hspace{-0.2cm} 83.95 & \hspace{-0.2cm} 77.86 & \hspace{-0.2cm} \textbf{79.20} & \hspace{-0.2cm} 77.73 & \hspace{-0.2cm} \textbf{80.13}\\

\hline
\end{tabular}

\end{table*}

\section{Results}

We conducted experiments using BERT and MuRIL-based models on four diverse datasets, and the corresponding results are presented in Table~\ref{Traning Accuracies}. The four models trained were IndicBERT, mBERT, MuRIL, and MahaBERT. The three-class classification was performed on four domain-specific datasets: MahaSent-PT (political tweets), MahaSent-MR (movie reviews), MahaSent-GT (general tweets), and MahaSent-ST (TV show subtitles). Additionally, we also evaluated the full mixed-domain corpus MahaSent-All. For the datasets MahaSent-PT, MahaSent-GT, and MahaSent-All, preprocessing involved the removal of hashtags, mentions, special symbols, and emoticons. However, MahaSent-ST and MahaSent-MR were clean datasets and did not require any specific preprocessing.

\subsection{MahaBERT, a monolingual Marathi BERT model works the best across datasets}
Upon analyzing the results, it was observed that the MahaBERT model achieved the highest accuracies for the MahaSent-PT, MahaSent-GT, and MahaSent-MR datasets compared to the MuRIL and other BERT-based models. Additionally, the Mahabert All model demonstrated an interesting ability to effectively learn from the shorter sentence structures present in TV show subtitles, resulting in the highest accuracy for the MahaSent-ST dataset.

Subsequently, we conducted further analysis on the performance of the best-performing model, MahaBERT, by comparing it across different domains. Initially, the model was trained on individual datasets, resulting in models named MahaBERT-PT, MahaBERT-GT, MahaBERT-ST, and MahaBERT-MR. Additionally, a model trained on the combined dataset was named MahaBERT-All. We then evaluated the accuracy scores of these models on the test sets of other domain datasets. The accuracy scores for all the models, with a maximum possible score of 100, are provided in Table~\ref{Cross-Domain Analysis}.

\subsection{Domain-specific models exhibit poor generalization, a mixed-domain model is more desirable}
Upon comparing the performance of different models, we observed that the MahaBERT-PT, MahaBERT-GT, and MahaBERT-MR models displayed impressive accuracy on their respective datasets. Interestingly, the MahaBERT-All model, trained on the combined dataset, exhibited the highest accuracy scores for both the MahaSent-ST and MahaSent-All datasets. This indicates the MahaBERT-All model's remarkable ability to generalize well across various datasets, highlighting its versatility and consistently achieving high or near the highest accuracy levels.

Additionally, we noticed that the domain-specific models did not perform well on out-domain test sets. This observation emphasizes the unique intricacies associated with each domain and underscores the importance of having a multi-domain dataset for comprehensive sentiment analysis. Out of the four domain-specific models MahaSent-ST trained on subtitles corpus gave the best cross-domain numbers. However, the numbers were significantly lower than the MahaBERT-All model.

\section{Conclusion}
In this paper, we present an enhanced version of the L3CubeMahaSent dataset, which expands beyond political tweets (MahaSent-PT) to include three additional domains: movie reviews (MahaSent-MR), general tweets (MahaSent-GT), and TV show subtitles (MahaSent-ST), each containing 15,000 examples. We fine-tuned four distinct models, namely MuRIL, mBERT, MahaBERT, and IndicBERT, on these datasets for evaluation purposes.

Our analysis revealed that the MahaBERT model consistently achieved the highest accuracies on the MahaSent-PT, MahaSent-GT, and MahaSent-MR datasets. To further investigate its performance across different domains, we conducted a cross-domain analysis using MahaBERT as the base model. Impressively, the MahaBERT-All model, trained on the combined dataset, demonstrated excellent generalization abilities, consistently achieving the highest or near-highest accuracies across diverse domains.

By providing a low-resource multi-domain dataset and models trained on specific domains, we aim to equip researchers and practitioners with valuable resources to analyze sentiment across diverse domains.

\section*{Acknowledgments}
This work was done under the L3Cube Pune mentorship
program. We would like to express our gratitude towards
our mentors at L3Cube for their continuous support and
encouragement. This work is a part of the L3Cube-MahaNLP project \cite{joshi2022l3cube_mahanlp}.

\nocite{langley00}

% \bibliography{icml2023/anthology,icml2023/example_paper}
\bibliography{main}
\bibliographystyle{icml2023}

%%%%%%%%%%%%%%%%%%%%%%%%%%%%%%%%%%%%%%%%%%%%%%%%%%%%%%%%%%%%%%%%%%%%%%%%%%%%%%%
%%%%%%%%%%%%%%%%%%%%%%%%%%%%%%%%%%%%%%%%%%%%%%%%%%%%%%%%%%%%%%%%%%%%%%%%%%%%%%%
% APPENDIX
%%%%%%%%%%%%%%%%%%%%%%%%%%%%%%%%%%%%%%%%%%%%%%%%%%%%%%%%%%%%%%%%%%%%%%%%%%%%%%%
%%%%%%%%%%%%%%%%%%%%%%%%%%%%%%%%%%%%%%%%%%%%%%%%%%%%%%%%%%%%%%%%%%%%%%%%%%%%%%%

\end{document}